\pdfoutput=1

\documentclass[11pt]{article}

\usepackage[final]{acl}

\usepackage{times}
\usepackage{latexsym}

\usepackage[T1]{fontenc}

\usepackage[utf8]{inputenc}

\usepackage{microtype}

\usepackage{inconsolata}

\usepackage{graphicx}

\usepackage{float}
\usepackage{adjustbox}

\usepackage{pifont}
\usepackage{soul}
\usepackage{bm}
\usepackage{hyperref}
\usepackage{seqsplit}
\usepackage{xcolor}
\usepackage{paralist}
\usepackage{amsmath, amsfonts}
\usepackage{algorithm,algpseudocode}
\algnewcommand\Input{\item[\algorithmicinput]}
\algnewcommand\algorithmicinput{\textbf{Input:}}
\algnewcommand\Output{\item[\algorithmicoutput]}
\algnewcommand\algorithmicoutput{\textbf{Output:}}
\usepackage{caption} 
\usepackage[capitalise,noabbrev]{cleveref}

\usepackage{booktabs}
\usepackage{array} 
\usepackage{multirow}
\usepackage{calc}
\usepackage{makecell}
\usepackage{listings}
\usepackage{spverbatim}
\usepackage{verbatimbox}
\usepackage{subfig}

\newcolumntype{P}[1]{>{\raggedright\arraybackslash}p{#1}}

\newcommand{\bs}{\textbackslash}
\newcommand{\ttt}[1]{\texttt{#1}}

\newsavebox{\verbboxone}

\title{CoAM: Corpus of All-Type Multiword Expressions}

\author{ \\
    \textbf{Yusuke Ide$^1$ \quad Joshua Tanner$^2$ \quad Adam Nohejl$^1$ \quad Jacob Hoffman$^2$ \quad Justin Vasselli$^1$} \\
    \textbf{Hidetaka Kamigaito$^{1}$ \quad Taro Watanabe$^1$}
    \\
    ${}^1$Nara Institute of Science and Technology \quad ${}^2$Resolve Research
    \\
    \normalsize \texttt{\{ide.yusuke.ja6, nohejl.adam.mt3, vasselli.justin\_ray.vk4, kamigaito.h, taro\}} \\
    \normalsize \texttt{@is.naist.jp, \{josh, jake\}@rslv.dev} \\
}

\begin{document}

\maketitle

\begin{abstract}
    Multiword expressions (MWEs) refer to idiomatic sequences of multiple words.
    MWE identification, i.e., detecting MWEs in text, can play a key role in downstream tasks such as machine translation, but existing datasets for the task are inconsistently annotated, limited to a single type of MWE, or limited in size.

    To enable reliable and comprehensive evaluation, we created CoAM: \textbf{C}orpus \textbf{o}f \textbf{A}ll-Type \textbf{M}ultiword Expressions, a dataset of 1.3K sentences constructed through a multi-step process to enhance data quality consisting of human annotation, human review, and automated consistency checking.
    Additionally, for the first time in a dataset for MWE identification, CoAM's MWEs are tagged with MWE types, such as \textsc{Noun} and \textsc{Verb}, enabling fine-grained error analysis.\footnote{
        The dataset is available at \url{https://huggingface.co/datasets/yusuke196/CoAM}.
    }
    Annotations for CoAM were collected using a new interface created with our interface generator, which allows easy and flexible annotation of MWEs in any form.\footnote{
        The source code of the interface generator is available at \url{https://github.com/Yusuke196/CAIGen}. 
    }
    Through experiments using CoAM, we find that a fine-tuned large language model outperforms MWEasWSD, which achieved the state-of-the-art performance on the DiMSUM dataset.
    Furthermore, analysis using our MWE type tagged data reveals that \textsc{Verb} MWEs are easier than \textsc{Noun} MWEs to identify across approaches.
\end{abstract}


\section{Introduction}
\label{sec:intro}

Vocabulary plays a critical role in the comprehension of natural language. 
While often simplified as a collection of single words, vocabulary also includes idiomatic sequences known as multiword expressions (MWEs) \cite{BaldwinMWETextbook}, which form an important part of language knowledge \citep{jackendoff_1995_boundaries}.
We focus specifically on sequences whose meaning or grammatical structure cannot be derived from their constituent words, as they can impede text analysis or comprehension; that is, we exclude transparent collocations.
For example, the bold sequences below constitute MWEs.
\begin{asparaenum}[(a)]
    \item \textit{ACL \textbf{stands for} Association for Computational Linguistics.}
    \item \textit{He has been \textbf{under the weather} lately.}
\end{asparaenum}

MWE identification (MWEI), the process of automatically tagging MWEs in text \citep{constant2017multiword}, is valuable for tasks where word-by-word text processing is insufficient, such as machine translation \cite{shuang2024} and lexical complexity assessment \cite{kochmar-etal-2020-detecting}.
\citet{briakou-etal-2024-translating} investigated step-by-step translation using a large language model (LLM) and demonstrated that identifying MWEs in the initial step improves the translation quality. 
Additionally, it is possible to build reading assistance systems to automatically gloss text by combining MWEI with definition modeling \citep{bevilacqua-etal-2020-generationary};
as \citet{huang-etal-2022-understanding} discusses, definition modeling will help laypeople comprehend specialized terminology, which includes many MWEs.

Despite the importance of MWEI, existing datasets for the task are inconsistently annotated, limited to a single type of MWE, or limited in size, as described in \cref{sec:related}.
This hinders reliable and comprehensive evaluations of MWEI systems.

In this paper, we introduce the \textbf{C}orpus \textbf{o}f \textbf{A}ll-type \textbf{M}ultiword expressions (CoAM), a dataset of 1.3K sentences for MWEI.
``Types'' refer to MWE categories assigned based on the part of speech of the MWE, with ``all-type'' signifying the inclusion of all such categories of MWEs.
To ensure annotation quality, we assigned two annotators and a reviewer to each sentence and checked all annotations for consistency.
Additionally, MWEs in the CoAM test set are tagged with their types, such as \textsc{Noun} and \textsc{Verb}, facilitating fine-grained error analysis of MWEI systems.
This enables us to address questions such as: ``Are verbal MWEs---the focus of the long-running PARSEME project \cite{savary-etal-2017-parseme}---more difficult to identify compared to other categories of MWEs?''

Annotations for CoAM were collected using a checkbox-based annotation interface created using our new interface generator, CAIGen: \textbf{C}heckbox-based \textbf{A}nnotation \textbf{I}nterface \textbf{Gen}erator.
It allows easy and flexible annotation of MWEs in any form, including discontinuous MWEs such as \textit{pick \ldots{} up} in \textit{\textbf{pick} me \textbf{up} at the station}, which are common in English and have historically been a challenge for MWEI systems \cite{rohanian-etal-2019-bridging}.

Using CoAM, we evaluate two distinct MWEI approaches.
The first approach, MWEasWSD (MaW, \citealp{tanner-hoffman-2023-mwe}), combines a rule-based pipeline and a trainable bi-encoder model, achieving state-of-the-art performance on the DiMSUM dataset \cite{schneider-etal-2016-semeval}.
The second is LLM fine-tuning for MWEI, inspired by the effectiveness of similar approaches in the task of named entity recognition (NER, \citealp{zhou2024universalner}).
We use CoAM to train and evaluate LLMs from the Llama \cite{dubey2024llama3herdmodels} and Qwen \cite{qwen2.5} model families.

The results reveal that the fine-tuned Qwen 72B model greatly outperforms MaW, demonstrating the effectiveness of LLM fine-tuning. 
On the other hand, all approaches suffer from low recall (e.g., 52.8\% for Qwen-72B).
Further analysis shows that fine-tuned LLMs struggle more with detecting \textsc{Noun} and \textsc{Clause} MWEs than with detecting \textsc{Verb} MWEs. 
MWEs not contained in WordNet \cite{Miller1995}, e.g., \textit{real estate}, were found particularly difficult to identify, presumably because they are less widely recognized as MWEs.


\section{Related Work}
\label{sec:related}

\paragraph{Datasets}
Previous studies have presented several datasets for MWEI and idiom\footnote{
    According to \citet{tedeschi-etal-2022-id10m}, idioms are a subset of MWEs.
} identification.
The \textbf{DiMSUM} \cite{schneider-etal-2016-semeval} dataset consists of 5,799 sentences annotated with MWEs, including, but not limited to, verbal MWEs, noun MWEs, phatics, and multi-word (MW) proper nouns.
Although DiMSUM has been used in multiple MWEI studies, e.g., \citet{kirilin-etal-2016-icl} and \citet{liu-etal-2021-lexical}, there exist inconsistencies in the annotation, which hinder proper evaluation.
\citet{tanner-hoffman-2023-mwe} found that over 80\% of the false positives of their system were actually caused by DiMSUM's inconsistent annotation.
Next, the \textbf{PARSEME} corpus \cite{savary-etal-2017-parseme, walsh-etal-2018-constructing} is a high-quality dataset for the identification of verbal MWEs.
It has been continuously updated, and the latest version (1.3) contains over 455,000 sentences across 26 languages. 
However, their focus is limited to verbal MWEs.
Lastly, \textbf{ID10M} \cite{tedeschi-etal-2022-id10m} is an idiom detection dataset consisting of automatically created training data in 10 languages and a manually curated evaluation benchmark of four languages. 
Their evaluation dataset was created with the help of professional annotators, but it contains only 200 sentences per language.
They used Wiktionary as the source of idioms, and MWEs not in Wiktionary were skipped in the annotation. 
Additionally, they did not annotate discontinuous idioms. 
Among non-English datasets for MWE identification, \textbf{Sequoia Corpus} \cite{candito2020french}, a French corpus annotated for MWEs and named entities, stands out for containing 3,440 MWEs. 

\paragraph{Tagging Schemes}
ID10M uses the \textbf{BIO} scheme. 
DiMSUM uses the more flexible \textbf{6-tag} scheme, which allows discontinuous MWEs, but cannot handle overlapping MWEs (see \cref{sec:appendix-overlapping} for an example of overlapping MWEs). 
In contrast, the \textbf{parseme-tsv} format \cite{savary-etal-2017-parseme} accepts MWEs in any form, which leads us to adopt an equivalent data format.

\paragraph{Other Tasks}
Whereas these studies addressed MWE/idiom identification as a sequence tagging task, others worked on a related but different task, idiom usage recognition.
It is a binary classification of word sequences in context as idiomatic/\allowbreak figurative or literal.
The detection task in \citet{muzny-zettlemoyer-2013-automatic}, SemEval-2013 Task 5b \cite{korkontzelos-etal-2013-semeval}, SemEval-2022 Task 2 Subtask A \cite{tayyar-madabushi-etal-2022-semeval}, and the MAGPIE corpus \cite{haagsma-etal-2020-magpie} are set up for this task.
Our task setting is more realistic and challenging than these, requiring systems to identify all MWEs within a given sentence.


\section{Task Formulation}
\label{sec:formulation}

We formulate the MWEI task as token-level sequence tagging, where the inputs are tokenized sentences and each token can belong to multiple MWEs.
Given the token sequence of each sentence, $x_1, \ldots{}, x_n$, the task is to output a list of MWEs where each MWE is represented as a list of token indices.
The $i$-th MWE in the sentence is represented with $[\mathrm{idx}(t_{i,1}), \mathrm{idx}(t_{i,2}), \ldots{}]$ where $t_{i,j}$ is the $j$-th token of the MWE and $\mathrm{idx}(t)$ is $t$'s index.
Our annotation interface allows annotations in this scheme, as illustrated in \cref{fig:interface}.




\section{Construction of CoAM}


\subsection{Data Selection and Preprocessing}
\label{sec:data-selection}

In selecting the sources of our dataset, we prioritized sources aimed at general audiences so that they are in standard English and mostly free of grammatical errors.
We also included both written text and transcribed speech.
Consequently, we utilized the following four data sources (see \cref{sec:data-sources} for their details). 
\textbf{News} is news text written by professional writers sourced from EMM NewsBrief, which was introduced by \citet{glavas-stajner-2013-event}.\footnote{
    The dataset also contains data from WikiNews and Wikipedia, but we preferred EMM NewsBrief, which contains more MWEs according to our preliminary analysis based on the annotation by \citet{kochmar-etal-2020-detecting}.
}
\textbf{Commentary} is commentaries on news from the WMT23 Shared Task monolingual training data \citep{kocmi-etal-2023-findings}. 
\textbf{TED} is a collection of TED talk transcriptions from two sources. 
    (1) \textbf{NAIST} is the dataset by \citet{Neubig2014-hv}; 
    (2) \textbf{IWSLT} is the subset of IWSLT 2017 Shared Task \cite{cettolo-etal-2017-overview}. 
\textbf{UD} is a collection of single sentences sourced from weblogs, reviews, question-answers, newsgroups, and emails in the English Web Treebank. 
It is part of the Universal Dependencies \citep{zeman-etal-2018-conll} and the English PARSEME corpus \cite{walsh-etal-2018-constructing}.
We did not include any sentences from UD in the test set, because UD contains user-generated content with frequent grammatical errors. 
All the other sources were used both for training and test splits.

For all sources, we took the first 10 (or all, when there are less than 10) sentences from each article or talk and presented them in original order. 
Each sentence was tokenized using spaCy (we use version \ttt{3.7.1} and the \ttt{en\_\allowbreak{}core\_\allowbreak{}web\_\allowbreak{}trf} model throughout this paper), with the exception of the UD sentences, which were already tokenized.
Note that we call tokens given in this manner \textit{words}. 


\subsection{Annotation Guidelines}

The construction of a reliable dataset requires clear guidelines for annotators in order to minimize misannotation.
However, we found no such guidelines for all-type MWE annotation, which led us to create new guidelines.

We define MWEs as idiomatic sequences that satisfy the following three conditions, based on the definition by the often-cited \citet{BaldwinMWETextbook} and the PARSEME annotation guidelines.\footnote{
    \url{https://parsemefr.lis-lab.fr/parseme-st-guidelines/1.3/} 
}

(a) \textbf{It consists of at least two words that are always realized by the same lexemes.}
This condition means that \textit{his} in \textit{put yourself in his shoes} is not part of the MWE, as other words like \textit{Michael's} can replace it.

(b) \textbf{It displays semantic, lexical, or syntactic idiomaticity.}
Semantic idiomaticity occurs when the meaning of an expression cannot be derived from its constituent words.
It is the most important type of idiomaticity because it accounts for the majority of MWEs being classified as idiomatic.
See additional discussion on semantic idiomaticity in relation to non-compositionality in \cref{sec:compositionality}.
Note that transparent collocations such as \textit{stuck at} are not MWEs in our definition because they are not semantically, lexically, or syntactically idiomatic.
The meaning or grammatical structure of transparent collocations can be understood by their constituents, and thus they can be processed word-by-word.

(c) \textbf{It is not a proper noun, i.e., a specific name of a person, facility, and so on.}
Previous MWE studies either classified MW proper nouns as MWEs \cite{schneider-etal-2016-semeval} or excluded them as non-idiomatic \cite{tayyar-madabushi-etal-2022-semeval}. 
We opted for the latter, as proper nouns are linked to \textit{encyclopedic} knowledge rather than \textit{lexicographic} knowledge \citep{NAVIGLI2012217}, to which typical MWEs belong to.


The full MWE definition is given in \cref{sec:guidelines}.
The guidelines were updated whenever an issue was found, e.g., an ambiguous description. 


\begin{figure*}[t!]
    \centering
    \includegraphics[width=1.0\linewidth]{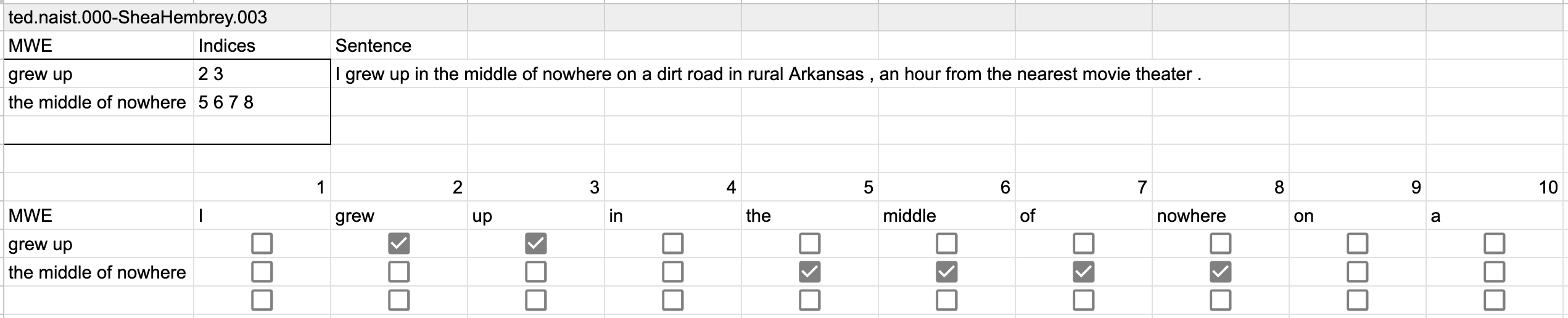} 
    \caption{
        An example sentence presented in our interface by CAIGen.
        Checks in the checkboxes are instantly reflected in the bordered zone at the upper right, showing which MWEs are currently marked. 
        Note that the number of rows is reduced here for brevity; the actual interfaces we used have nine rows per sentence.
    }
    \label{fig:interface}
\end{figure*}


\subsection{Annotation Interface}

For flexible and efficient annotation, we developed a novel annotation interface generator, CAIGen.
CAIGen builds a checkbox-based interface in Google Sheets\footnote{
    \url{https://developers.google.com/sheets}
}, allowing annotators to perform annotations of MWEs and other kinds of spans simply by checking checkboxes as shown in \cref{fig:interface}.
In the interface generation process, for each sentence, CAIGen first writes the sentence ID, the sentence itself, and a bordered box to show the result of annotations. 
Then, it arranges each token from left to right, wrapping the line when its length hits a pre-set limit.

\begin{table}[t!]
    \centering
    \small
    \begin{tabular}{
            >{\hspace{-3pt}}l
            >{\hspace{-4pt}}c
            >{\hspace{-4pt}}c
            >{\hspace{-4pt}}c
            >{\hspace{-4pt}}c<{\hspace{-3pt}}
        }
        \toprule
         & 
            \makecell{Flexible\\annota-\\tion} & 
            \makecell{Easy\\collabora-\\tion} & 
            \makecell{Easy-to-use\\interface} & 
            \makecell{Customiz-\\able\\interface} \\

        \midrule
        brat & \color{lightgray}\ding{55} & \color{lightgray}\ding{55} & \ding{51} & \color{lightgray}\ding{55} \\
        FLAT & \ding{51} & \color{lightgray}\ding{55} & \color{lightgray}\ding{55} & \color{lightgray}\ding{55}\\
        \textbf{CAIGen} & \ding{51} & \ding{51} & \ding{51} & \ding{51} \\
        \bottomrule
    \end{tabular}
    \caption{
        Comparison of annotation tools.
    }
    \label{tab:annotation_tools}
\end{table}

Its main advantages over other comparable annotation tools are summarized in \cref{tab:annotation_tools}.
The first tool, \textbf{brat}\footnote{
    \url{https://github.com/nlplab/brat}
} \cite{stenetorp2012brat}, has been used for annotation of a wide range of tasks such as sentiment analysis \citep{Pontiki2016SemEval2016T5} and NER \citep{Tabassum2020CodeAN}. 
\textbf{FLAT}\footnote{
    \url{https://github.com/proycon/flat}
} was used for the annotation of the PARSEME corpus \cite{savary-etal-2017-parseme}.
The advantages of CAIGen over these are as follows.
First, the generated interface supports annotations of spans of any form, including discontinuous or overlapping spans (see \cref{sec:appendix-notes} for examples), making it more flexible than the previous tools.
Second, CAIGen alleviates researchers' overhead of managing servers and annotator accounts by delegating to Google, while the other tools require either (1) distributing data slices to each annotator and having them run applications locally or (2) running and managing the application on a remote server.
Third, CAIGen provides an easy-to-use spreadsheet-based interface, which is simple and familiar to technical and non-technical annotators alike.
Lastly, our interface is highly customizable because CAIGen builds it using Google Apps Script, a well-documented programming language based on JavaScript; for example, additional annotations of span types could be easily collected. 


\subsection{Annotation}

The annotation was done by two annotators hired through a company and five authors.
The hired annotators comprise one native and one non-native English speaker, both with at least six years of experience as translators.
They received feedback from the authors after annotating the first 50 or so sentences to correct misunderstandings about the guidelines. 
We paid 154 JPY (roughly 1 USD) per sentence as a reward to the company.
Meanwhile, the author annotators consist of three native and two non-native speakers, all with a language-related degree and sufficient English proficiency to perform the task.
For reliable annotation, we assigned each sentence to two annotators (one hired annotator and one author), ensuring that at least one was a native English speaker, as the task requires a rich English vocabulary.

All annotators were instructed to carefully read the guidelines and perform annotation using the checkbox-based interface.
In the authors' annotation, we marked unclear sentences to remove them later because it is hard to determine whether a span is an MWE in such sentences. 


\subsection{Inter-Annotator Agreement}

With two annotators being assigned to each sentence, we measured the inter-annotator agreement (IAA) by the MWE-based (exact-match) F1 score.
The IAA was found to be only 37.3\%, which could be attributed to annotators failing to annotate valid MWEs.
We thus suggest that MWE datasets be constructed with at least two independent annotators to reduce missed MWEs and that the annotators be adequately trained to ensure they fully understand the annotation guidelines.

Nonetheless, it should be noted that this IAA is comparable with other MWEI datasets, and that it does not reflect the quality of CoAM.
Our IAA is computed in a strict way, which we chose for its clear interpretability.
CoAM's IAA is comparable to those of other datasets when the same, more lenient methods are used.
PARSEME 1.1 \cite{ramisch-etal-2018-edition} computes F1 for exact span matches as we do, but they only report ``the highest scores among all possible annotator pairs'' (see Table~1, ibid.).
The highest score for English PARSEME is 52.9\%, similar to the highest score computed in the same way for CoAM: 52.2\%.
Meanwhile, DiMSUM was annotated by a single annotator, and thus no actual agreement was reported.
Although \citet{schneider-etal-2016-semeval} estimates DiMSUM's IAA to range from 60\% to 75\%, this is based on a very small subset (66 sentences) reannotated by the first author and computed as a more lenient F1 score based on partial span matches.
Thus, while our reported IAA value is low, we emphasize that our IAA is in line with comparable MWEI datasets.
Additionally, we addressed the annotation disagreements through the combination of human review and consistency checking as described below.


\begin{table*}[th!]
    \centering
    \small
    \begin{tabular}{
        >{\hspace{-2pt}}
        p{0.07\linewidth}<{\hspace{-2pt}}
        P{0.14\linewidth}<{\hspace{-2pt}}
        >{\raggedright\arraybackslash}p{0.42\linewidth}<{\hspace{-2pt}}
        >{\raggedright\arraybackslash}p{0.25\linewidth}<{\hspace{-2pt}}
    }
        \toprule
        Type & PoS of Head & Description & Example \\

        \midrule
        \multirow[c]{2}{*}{\textsc{Noun}} & NOUN, PRON, PROPN & Noun MWEs or compounds. & \textit{the \ul{middle} of nowhere, red \ul{tape}} \\

        \cmidrule(r){1-1} \cmidrule(l){2-4}
        \textsc{Verb} & VERB, AUX & Verbal MWEs. & \textit{\ul{stand} for, \ul{pick} up, \ul{be} the case} \\

        \cmidrule(r){1-1} \cmidrule(l){2-4}
        \makecell[lt]{\textsc{Mod}/\\\textsc{Conn}} & ADJ, ADV, ADP, CCONJ, SCONJ & Adjectival, adverbial, adpositional, or connective MWEs. & \textit{\ul{under} the weather, \ul{of} course, \ul{in} spite of} \\

        \cmidrule(r){1-1} \cmidrule(l){2-4}
        \multirow[c]{2}{*}{\textsc{Clause}} & VERB, AUX & MWEs containing (1) a verb or auxiliary verb and (2) its nominal subject, such as phatics and proverbs. & \textit{you\ul{'re} welcome, the early bird \ul{gets} the worm, when it \ul{comes} to} \\

        \cmidrule(r){1-1} \cmidrule(l){2-4}
        \multirow[c]{2}{*}{\textsc{Other}} & Any & MWEs whose head is not contained in them or whose PoS is none of the PoS above. & \textit{and so on, \ul{oh} man} \\
        \bottomrule
    \end{tabular}
\caption{MWE types in CoAM. The PoS are denoted with UPOS tags. Underlines denote the head of an MWE.}
\label{tab:mwe_types}
\end{table*}


\subsection{Review}
\label{sec:review}

To solve the disagreement between the two annotators and correct any other problematic annotations, two authors---both native English speakers---reviewed all the annotations.
We presented them with the annotations in a special interface for reviewing, where the tags given by only one annotator were highlighted, and we asked them to mark inappropriate tags to be deleted and newly found MWEs to be added.
When the review was complete, we updated our dataset according to the marks added by reviewers.


\begin{table}[t!]
    \centering
    \small
    \begin{tabular}{p{5cm}c} 
        \toprule
        Sentence & Marked \\
        \midrule
        Having never booked train tickets online before thought I would \textbf{give} it a \textbf{try} and was very surprised at how much I saved. & \ding{51} \\
        \cmidrule(r){1-1} \cmidrule(l){2-2}
        Would recomend \textbf{giving} this a \textbf{try}. & \color{lightgray}\ding{55} \\
        \bottomrule
    \end{tabular}
    \caption{Inconsistent annotation of \texttt{give\_try}. Note that the second sentence is reproduced as in DiMSUM, including apparent typos.}
    \label{tab:inconsistent_annotation}
\end{table}


\subsection{Consistency Check}
\label{sec:consistency-check}

One of the primary issues with the annotations in the DiMSUM dataset is their inconsistency---that is, a number of MWEs are annotated in one location and not another, despite equivalent constituents and semantics being present in both places. For example, see how \texttt{give\_try} is labeled in only one of the sentences in \cref{tab:inconsistent_annotation}.

In order to quantify this issue in both DiMSUM and CoAM---and to eliminate it from CoAM---we used a partially automated approach to find inconsistencies in both the entirety of CoAM and 1.3K sentences (the size of CoAM) randomly sampled from DiMSUM. We started by initializing an empty set $M$, then iterated through all given sentences and added all labeled MWEs from them to $M$. Next, we used a simple rule-based MWEI pipeline, repurposed from \citet{tanner-hoffman-2023-mwe}, to find constituent groups in other sentences that could correspond to an MWE in $M$ but were not already labeled. Finally, an author and native speaker of English reviewed each of these candidate constituent groups to see if they are semantically equivalent to already labeled instances of this MWE---that is, to see if they represent an inconsistency. 

We found 118 instances of inconsistencies like this in the random sample of DiMSUM and 147 in CoAM,\footnote{
    When counting inconsistencies, we always consider positive labels correct, and negative labels inconsistent.
} reaffirming the difficulty involved in producing consistent MWE annotations.
However, we then added the missing labels to all MWEs found in this way, eliminating these inconsistencies from the final CoAM data.


\begin{table*}[th!]
    \centering
    \small
    \begin{tabular}{lccccccccc}
    \toprule
     & & & & & \multicolumn{5}{c}{MWE Type Proportion (\%)} \\
    \cmidrule(l){6-10}
     & Sentences & Words & MWEs & \makecell{MWE\\Density (\%)} & \textsc{Noun} & \textsc{Verb} & \textsc{Mod/Conn} & \textsc{Clause} & \textsc{Other} \\
    \midrule
    News & 360 & 9,328 & 230 & 5.5 & 32.2 & 44.3 & 23.0 & 0.0 & 0.4 \\
    Commentary & 357 & 9,310 & 269 & 7.0 & 30.5 & 29.7 & 33.1 & 1.5 & 5.2 \\
    TED & 299 & 6,592 & 210 & 7.1 & 35.2 & 38.6 & 21.4 & 3.8 & 1.0 \\
    UD & 285 & 5,001 & 158 & 7.2 & 42.4 & 39.2 & 12.7 & 1.3 & 4.4 \\
    \midrule
    Train & 780 & 16,817 & 486 & 6.7 & 36.2 & 38.3 & 19.8 & 1.4 & 4.3 \\
    Test & 521 & 13,414 & 381 & 6.4 & 31.8 & 36.5 & 29.1 & 1.8 & 0.8 \\
    \midrule
    Total & 1,301 & 30,231 & 867 & 6.6 & 34.3 & 37.5 & 23.9 & 1.6 & 2.8 \\
    \bottomrule
    \end{tabular}
    \caption{Statistics of CoAM. MWE density is the percentage of words in MWEs. The test set comprises part of News, Commentary, and TED, while the training set comprises the rest of the data.}
    \label{tab:stats}
\end{table*}


\subsection{MWE Type Tagging}
\label{sec:mwe-type-tagging}

To enable fine-grained error analysis, we automatically tagged all the MWEs in CoAM with MWE types. 
We then manually corrected the tags in the test set, as described below, to enable a precise analysis of the results by MWE types.

Inspired by the classification by \citet{schneider2014comprehensive}, we group MWEs into five types: \textsc{Noun}, \textsc{Verb}, \textsc{Modifier/Connective} (\textsc{Mod/Conn}),\footnote{
    We include adjectival, adverbial, adpositional, and connective MWEs in the category of \textsc{Mod/Conn} to create a category with a sufficient data size.
} \textsc{Clause}, and \textsc{Other}.
The details are described in \cref{tab:mwe_types}.
We automatically tag MWE types in each sentence based on the dependency structure.
Dependency parses were provided in the data source for UD and obtained using spaCy for News, Commentary, and TED.
For each MWE $\bm{m} = (w_1,\ldots,w_{|\bm{m}|})$, we look for its syntactic head, namely, try to find $w^*$ that has all the other words in $\bm{m}$ as its descendants.
If $\bm{m}$ has such $w^*$, we determine its type based on the PoS of $w^*$; for example, we tag $\bm{m}$ as \textsc{Verb} when $w^*$ is a verb.
If $\bm{m}$ does not have such $w^*$, we tag $\bm{m}$ as \textsc{Other}.
For the details, see the algorithm in \cref{sec:appendix-type-tagging}.

We then manually checked all 384 MWEs in the test set and corrected the errors in them.
Two authors who are native English speakers conducted this, identifying and correcting 30 errors. The error rate of automatic tagging was therefore 7.8\%. The most common errors were \textsc{Mod/Conn} automatically tagged as \textsc{Verb} (e.g., \textit{according to}) and \textsc{Other} (e.g., \textit{not only}) constituting 3.3\% and 3.1\% of the test data, respectively. We expect this to be representative of the errors in the type tags automatically assigned to the training set. Note, however, that we used only the manually corrected tags on the test set for result analysis in \cref{sec:results}.


\subsection{Statistics}

\cref{tab:stats} shows dataset statistics.
CoAM has more than 1.3K sentences.
The MWE density is 6--7\% in both training and test sets.
This proportion is much lower than that of DiMSUM \cite{schneider-etal-2016-semeval}---13\% overall---most likely because DiMSUM includes proper noun phrases in MWEs.
Among the five MWE types, \textsc{Verb} and \textsc{Noun} are the most frequent ones across data sources.
See \cref{sec:appendix-stats} for additional analysis regarding MWE continuity.



\section{MWEI Approaches}

We use CoAM to evaluate two MWEI approaches.


\subsection{MWEasWSD}

MWEasWSD (MaW) is an approach that uses (1) an MWE lexicon---WordNet \cite{Miller1995}---and a rule-based pipeline to identify MWE candidates and (2) a trainable model to filter MWE candidates \cite{tanner-hoffman-2023-mwe}.
It achieved state-of-the-art performance on the DiMSUM dataset.
They published four of their filtering models, among which we use the bi-encoder (BiEnc) and DCA poly-encoder (DCA).
Both models are based on BERT \cite{devlin-etal-2019-bert}, specifically \ttt{bert-base-uncased}.
They have been trained with SemCor \cite{miller-etal-1993-semantic}, and we further fine-tune each model with the CoAM training set.
We also run MaW with the rule-based pipeline only, i.e., without a filtering model.


\subsection{LLM Fine-Tuning}
\label{sec:inst-tuning}

Recent studies have shown the effectiveness of LLM fine-tuning for a wide range of tasks, such as NER \cite{zhou2024universalner} and grammatical error correction \cite{kaneko-okazaki-2023-reducing}. 
Framing their task as sequence transduction, they achieved high performance by providing LLMs with prompts that included instructions on the task.

Inspired by their success, we perform LLM fine-tuning, where the inputs to LLMs are instructions for MWEI (a summary of the annotation guidelines) followed by formatted tokens in a sentence.


\section{Experiments}

\subsection{Setup}


\paragraph{LLM Fine-Tuning}
We use four instruction-tuned LLMs that are available on Hugging Face Hub: \ttt{Llama-\allowbreak{}3.1-\allowbreak{}8B-\allowbreak{}Instruct}, \ttt{Llama-\allowbreak{}3.1-\allowbreak{}70B-\allowbreak{}Instruct} \cite{dubey2024llama3herdmodels}, \ttt{Qwen-\allowbreak{}2.5-\allowbreak{}7B-\allowbreak{}Instruct}, and \ttt{Qwen-\allowbreak{}2.5-\allowbreak{}72B-\allowbreak{}Instruct} \cite{qwen2.5}.
We abbreviate them, e.g., to Llama-8B, omitting the versions.
For efficient training and inference, we use QLoRA \cite{dettmers2023}, performing 4-bit NormalFloat quantization and double quantization. 
Other hyperparameters are described in \cref{sec:setup-it-appendix}.
The computational budgets are in \cref{sec:appendix-budgets}.

To find a performant input-output format, we conduct preliminary experiments with three formats, adjusting the prompt for each format.
These experiments are detailed in \cref{sec:setup-it-appendix}.
We find \verb|tsv_to_tsv| (see \cref{tab:it-prompt}) is the only format the models can comply with and thus adopt it.

All prompts contain a placeholder for a definition of MWEs to clarify what sequences should be marked.
We report scores using what we refer to as the \textit{long definition} in \cref{tab:results}.
For scores using a shorter definition, see the ablation study in \cref{para:mwe-def}.


\begin{table}[t!]
    \centering
    \small
    \begin{tabular}{
        >{\hspace{-2pt}}l
        >{\hspace{-4pt}}p{0.76\linewidth}
        <{\hspace{-2pt}}
    }
        \toprule
        Role & Message \\
        \midrule
        \multirow[c]{-2}{*}{System} & \begin{verbbox}
You are a helpful system to identify
multiple-word expressions (MWEs).
\end{verbbox} 
            \theverbbox \\
        \cmidrule(r){1-1} \cmidrule(){2-2}
        \multirow[c]{-18}{*}{User} & \begin{verbbox}
Identify all the MWEs in the given 
sentence, and output their surface 
forms and the indices of their 
components.\n
\n
[MWE_DEFINITION]\n
\n
Each sentence is given as a string of 
words delimited by '\n'. Respond in 
TSV format, where the first and second 
columns contain words and MWE tags, 
respectively. The MWE tag should be a 
string of MWE identifiers. When a word 
belongs to multiple MWEs, the tag 
should be a concatenation of their 
numbers delimited by semicolons.\n
\n
Sentence:\n
ACL\n
stands\n
\end{verbbox}
            \theverbbox \\
            & \ttt{\quad\vdots} \\

        \bottomrule
    \end{tabular}
    \caption{
        Example prompt for fine-tuning, based on the \ttt{tsv\_to\_tsv} format.
        In our main experiments, \ttt{[MWE\_\allowbreak{}DEFINITION]} will be filled with the long MWE definition described in \cref{para:mwe-def}.
    }
    \label{tab:it-prompt}
\end{table}


\paragraph{Evaluation Metrics}
We use MWE-based precision, recall, and F1 score \cite{savary-etal-2023-parseme}.
See \cref{sec:metrics} for their exact definitions.


\begin{table*}[th!]
    \centering
    \small

    \subfloat[\label{tab:results_a}]{
        \begin{tabular}{llccccccc}
        \toprule
         & & & & & \multicolumn{4}{c}{Recall by MWE Type} \\
        \cmidrule(l){6-9}
         & & F1 & P & R & \textsc{Noun} & \textsc{Verb} & \textsc{Mod/Conn} & \textsc{Clause} \\
         &  &  &  &  & (121) & (139) & (111) & (7) \\

        \midrule
        \multirow[c]{4}{*}{FT} & Llama-8B & 24.9$_{\pm 3.7}$ & $\mathbf{92.0}$$_{\pm 2.0}$ & 14.4$_{\pm 2.4}$ & 3.3$_{\pm 0.0}$ & 19.4$_{\pm 2.9}$ & 21.6$_{\pm 4.8}$ & 0.0$_{\pm 0.0}$ \\
         & Llama-70B & 38.3$_{\pm 5.2}$ & 69.0$_{\pm 1.7}$ & 26.8$_{\pm 5.1}$ & 19.0$_{\pm 5.2}$ & 39.6$_{\pm 7.5}$ & 21.6$_{\pm 2.7}$ & 0.0$_{\pm 0.0}$ \\
         & Qwen-7B & 48.1$_{\pm 1.0}$ & 60.9$_{\pm 0.8}$ & 39.7$_{\pm 1.0}$ & 28.7$_{\pm 1.3}$ & 52.0$_{\pm 0.8}$ & 39.6$_{\pm 0.9}$ & 0.0$_{\pm 0.0}$ \\
         & Qwen-72B & $\mathbf{57.8}$$_{\pm 1.8}$ & 63.8$_{\pm 2.2}$ & $\mathbf{52.8}$$_{\pm 2.1}$ & $\mathbf{42.1}$$_{\pm 4.4}$ & $\mathbf{59.7}$$_{\pm 1.4}$ & $\mathbf{57.7}$$_{\pm 1.8}$ & $\mathbf{28.6}$$_{\pm 0.0}$ \\

        \midrule
        \multirow[c]{3}{*}{MaW} & Rule & 32.4 & 27.9 & 38.6 & 33.1 & 45.3 & 39.6 & 0.0 \\
         & Rule+BiEnc & 41.4$_{\pm 0.3}$ & 47.9$_{\pm 0.8}$ & 36.5$_{\pm 0.3}$ & 28.9$_{\pm 0.0}$ & 45.3$_{\pm 0.0}$ & 36.9$_{\pm 0.9}$ & 0.0$_{\pm 0.0}$ \\
         & Rule+DCA & 41.9$_{\pm 0.2}$ & 49.0$_{\pm 0.5}$ & 36.7$_{\pm 0.5}$ & 29.8$_{\pm 0.0}$ & 45.3$_{\pm 0.7}$ & 36.6$_{\pm 1.0}$ & 0.0$_{\pm 0.0}$ \\

        \bottomrule
        \end{tabular}
    }

    \subfloat[\label{tab:results_b}]{
        \begin{tabular}{
            ll
            c<{\hspace{4pt}} 
            c
            c
            c
            c<{\hspace{4pt}} 
            c
        }
        \toprule
         & & \multicolumn{2}{c}{Recall by Seen/Unseen} & \multicolumn{2}{c}{Recall by Continuity} & \multicolumn{2}{c}{Recall by In WN or Not} \\
        \cmidrule(lr){3-4} \cmidrule(lr){5-6} \cmidrule(l){7-8}
         &  & Seen & Unseen & Continuous & Discontinuous & True & False \\
         &  & (141) & (240) & (338) & (43) & (161) & (220) \\

        \midrule
        \multirow[c]{4}{*}{FT} & Llama-8B & 29.6$_{\pm 5.0}$ & 5.6$_{\pm 0.9}$ & 16.0$_{\pm 2.7}$ & 2.3$_{\pm 0.0}$ & 23.4$_{\pm 3.2}$ & 7.9$_{\pm 1.8}$ \\
         & Llama-70B & 37.1$_{\pm 4.5}$ & 20.7$_{\pm 5.4}$ & 29.0$_{\pm 4.9}$ & 9.3$_{\pm 6.2}$ & 43.9$_{\pm 6.3}$ & 14.2$_{\pm 4.1}$ \\
         & Qwen-7B & 47.3$_{\pm 0.4}$ & 35.3$_{\pm 1.3}$ & 43.9$_{\pm 1.1}$ & 7.0$_{\pm 0.0}$ & 53.8$_{\pm 0.9}$ & 29.4$_{\pm 1.1}$ \\
         & Qwen-72B & $\mathbf{61.9}$$_{\pm 0.8}$ & $\mathbf{47.4}$$_{\pm 3.8}$ & $\mathbf{57.3}$$_{\pm 1.8}$ & $\mathbf{17.1}$$_{\pm 4.8}$ & 62.7$_{\pm 2.8}$ & $\mathbf{45.5}$$_{\pm 2.8}$ \\

        \midrule
        \multirow[c]{3}{*}{MaW} & Rule & 46.8 & 33.8 & 41.7 & 14.0 & $\mathbf{91.3}$ & 0.0 \\
         & Rule+BiEnc & 45.2$_{\pm 0.4}$ & 31.4$_{\pm 0.2}$ & 39.6$_{\pm 0.3}$ & 11.6$_{\pm 0.0}$ & 86.3$_{\pm 0.6}$ & 0.0$_{\pm 0.0}$ \\
         & Rule+DCA & 44.0$_{\pm 1.2}$ & 32.4$_{\pm 0.2}$ & 40.3$_{\pm 0.5}$ & 7.8$_{\pm 1.3}$ & 86.7$_{\pm 1.3}$ & 0.0$_{\pm 0.0}$ \\
        \bottomrule
        \end{tabular}
    }
    \caption{
        Results by MWEI system and metric/MWE category, as mean percentage scores of three runs with random training seeds.  
        The numbers in parentheses are the MWE counts.
        $\pm$ denotes standard deviation.
        The bold font denotes the highest score.
        \textit{Rule} stands for the rule-based baseline, and \textit{WN} for WordNet.
    } 
    \label{tab:results}
\end{table*}


\begin{table*}[t!]
    \centering
    \small
    \begin{tabular}{
        >{\hspace{-2pt}}
        l<{\hspace{-2pt}}
        l<{\hspace{-2pt}}
        l<{\hspace{-2pt}}
        l<{\hspace{-2pt}}
    }
        \toprule
        Result & MWE & Context & Note \\

        \midrule
        TP & \textit{fire up} & \textit{The allegations have \textbf{fired up} the opposition, \ldots{}} & \textsc{Verb} in WordNet \\
        TP & \textit{at least} & \textit{\ldots{} since \textbf{at least} the 1950s.} & \textsc{Mod/Conn} in WordNet \\
        FN & \textit{in \ldots hands} & \textit{\ldots concentration of power \textbf{in} his own \textbf{hands}.} & \textsc{Mod/Conn} in WordNet \\
        FN & \textit{real estate} & \textit{\ldots park their toxic \textbf{real estate} assets \ldots} & \textsc{Noun} not in WordNet \\
        FN & \textit{you know} & \textit{\textbf{You know}, it's very old \ldots{}} & \textsc{Clause} not in WordNet \\
        \bottomrule
    \end{tabular}
    \caption{
        Examples of true positives (TPs) and false negatives (FNs), i.e., MWEs identified/missed in all three runs, of fine-tuned Qwen-72B.
    }
    \label{tab:tp_and_fn}
\end{table*}


\subsection{Results and Analysis}
\label{sec:results}

In \cref{tab:results_a}, the left three columns show the overall scores.
Fine-tuned Qwen-72B achieves the best F1, surpassing the highest F1 by Rule+DCA of MaW.
Additionally, fine-tuned Qwen-72B outperforms all MaW systems in precision and recall.
This indicates the effectiveness of fine-tuning LLMs for MWEI---particularly LLMs with a large number of parameters.
One explanation for this could be that knowledge about MWEs was acquired by LLMs through pre-training, and we can harness this knowledge for MWEI by fine-tuning.

Meanwhile, all systems, except for the rule-based one, suffer from low recall.
Even the best system, fine-tuned Qwen-72B, achieves a recall of only 52.8\%, missing almost half of the gold MWEs. 


\paragraph{Analysis of Recall}
Given the low recall across all systems, we conduct further analysis to examine which MWEs are particularly difficult to identify.

Recall by MWE Type in \cref{tab:results_a} shows that \textsc{Clause} and \textsc{Noun} MWEs tend to be more difficult to identify than \textsc{Mod/Conn} or \textsc{Verb} MWEs across models.

In \cref{tab:results_b}, the first two columns analyze how much recall changes depending on whether the MWE is seen or unseen, where an MWE in the test set is considered unseen if the multi-set of lemmas of its constituents was never annotated in the training set.
We find that seen MWEs are easier to identify than unseen ones across models and systems, which resembles the results of the PARSEME shared task 1.2 \cite{ramisch-etal-2020-edition}.

Recall by Continuity in \cref{tab:results_b} shows that discontinuous MWEs are, across models and systems, far more difficult to identify than continuous ones, indicating that discontinuous MWEs remain a key challenge and require further attention.

Recall by In WN or Not in \cref{tab:results_b} reveals that MWEs not contained in WordNet are much more difficult to identify than those in WordNet.
It is natural that MaW systems cannot identify the MWEs not in WordNet, as their rule-based pipeline cannot detect these candidates, but interestingly we find that fine-tuned LLMs also struggle to identify MWEs not in WordNet.
We hypothesize that this is caused by MWEs in WordNet being more widely recognized as MWEs or idioms and that this recognition is reflected in the training data of the LLMs, enhancing their ability to identify these MWEs.

\cref{tab:tp_and_fn} shows examples of correctly identified MWEs and missed MWEs by fine-tuned Qwen-72B.
\textsc{Verb} MWEs in WordNet like \textit{fire up} are relatively easy to identify, with a recall of 79.0\% achieved by the model.
Meanwhile, the discontinuous MWE, \textit{in \ldots hands}, was not correctly identified.
Additionally, the \textsc{Noun} MWE \textit{real estate} was not identified; the aforementioned hypothesis could explain this, as \textit{real estate} is not in WordNet.
Accurately identifying those MWEs could be a challenge for future studies.


\begin{table*}[th!]
    \centering
    \small

    \begin{tabular}{
        l
        ccc
        cc
        >{\hspace{14pt}}cc
    }
    \toprule
         & & & & \multicolumn{2}{c}{Recall by Continuity} & \multicolumn{2}{c}{Recall by In WN or Not} \\
    \cmidrule(lr){5-6} \cmidrule(lr){7-8}
         & F1 & P & R & Continuous & Discontinuous & True & False \\
    \midrule
    Human & 56.5 & 64.5 & 51.0 & 52.1 & 41.8 & 56.8 & 47.0 \\
    \bottomrule
    \end{tabular}

    \caption{Mean human agreement with the final CoAM annotations. Both train and test sets are included in the data.}
    \label{tab:results_human}
\end{table*}


\begin{table}[t!]
    \centering
    \small
    \begin{tabular}{lcc}
    \toprule
     & Llama-70B & Qwen-72B \\

    \midrule
    FT & 38.3$_{\pm 5.2}$ & 57.8$_{\pm 1.8}$ \\
    \midrule
    FSL & 6.1$_{\pm 0.0}$ & 12.9$_{\pm 0.0}$ \\
    ZSL & 3.9 & 1.4 \\
    \bottomrule
    \end{tabular}

    \caption{
        Ablation results by MWEI method and model, as mean percentage F1 scores of three runs (for FT the randomness arises from training, and for FSL from example selection).
        $\pm$ denotes standard deviation.
    }
    \label{tab:ablation}
\end{table}

\paragraph{Ablation Study on LLM Fine-Tuning}
Given the relatively high performance of fine-tuned Qwen-72B, we investigate how much fine-tuning (FT) contributes to the performance by comparing FT to zero-shot learning (ZSL) and few-shot learning (FSL).
In ZSL, we provide the models with the same prompt as FT.
In FSL, we sample 5 pairs of input (sentence) and output (gold MWE set) from the training set, convert them into the \verb|tsv_to_tsv| format, and include them in the prompt as examples.
In the sampling process, we ensure that the models have a sufficient number of examples to learn the task and format by repeating random sampling until at least two of the example sentences contain one or more MWEs.

\cref{tab:ablation} shows the results. 
FT greatly outperforms ZSL and FSL, demonstrating the effectiveness of fine-tuning on the CoAM training set.


\paragraph{Comparison considering computational efficiency}

As shown in \cref{tab:results_a}, fine-tuned Qwen-72B outperforms the best MaW system in all of the F1 score, precision, and recall.
However, MaW is more efficient than the LLM in terms of memory and compute.
MaW's encoder model (\ttt{bert-\allowbreak{}base-\allowbreak{}uncased}) has only 110M parameters, taking approximately $2 \times 110\textrm{M} \times 4 = 880\textrm{M}$ bytes in total ($2$ is the number of encoders).
4-bit quantized Qwen-72B takes $72,000\textrm{M} \times 0.5 = 36,000\textrm{M}$ bytes, taking roughly 40 times more space than MaW.
On the other hand, the performance of 4-bit quantized Qwen-7B, which takes roughly 4 times more space, is comparable with Rule+DCA.


\section{Discussion}

How do the errors made by the automated approaches contrast with human annotators' errors?
To address this question, we perform an analysis using the final CoAM annotations as gold annotations and the raw human annotations as predictions.
This is not a completely fair setting for evaluating the humans' capability of MWEI, as the human annotations partially contribute to the gold standard;
however, annotations by an annotator are only one factor in determining the final gold annotations.
The gold annotations on each sentence are also affected by another annotator, a reviewer, and the consistency checking process. 
We thus expect this analysis to shed light on the characteristics of MWEI performance for humans and the systems.

\cref{tab:results_human} shows the human agreement with the gold annotations.
Here we find all of F1, precision, and recall are similar to those of fine-tuned Qwen-72B; on average, human recall is only around 50\%.
However, closer analysis reveals this is not true for all the annotators.
The top annotator achieved an F1 of 72.0\% with a recall of 71.9\%, while the runner-up annotator reached an F1 of 66.0\% with a recall of 59.4\%.\footnote{
    It should be noted that the annotators worked on different sentences, but we assume the annotation difficulty varied minimally.
}
The top annotator holds a linguistics-related degree, and the second has authored a paper on MWEI, which may have served as relevant training for MWEI annotation.
These results suggest that automated systems could also achieve higher MWEI performance if the underlying models were better trained.

In \cref{sec:results}, we discussed that discontinuous MWEs remain a key challenge for the MWEI systems.
In contrast, Recall by Continuity in \cref{tab:results_human} shows that discontinuous MWEs are not especially challenging for humans, compared to continuous MWEs.
This indicates that better modeling for discontinuous MWEs could greatly improve the systems' performance on this category of MWEs.


\section{Conclusion}

In this paper, we constructed CoAM, a high-quality dataset of 1.3K sentences for MWEI.
Using a combination of human review and automated consistency checking, we addressed consistency issues that have been a problem for previous MWEI datasets, enabling more accurate evaluation results.
We used CoAM to evaluate two MWEI approaches: MaW and LLM fine-tuning.
Our largest fine-tuned LLM performed the best, but all systems suffered from low recall.
Consequently, we argue that MWEI overall remains a challenging task.


\section*{Limitations}
\label{sec:limitations}

\paragraph{Dataset Size}
Because the objective of this work was to maximize dataset quality, we invested heavily in efforts to improve data quality, employing multiple annotators per sentence, manual review, and consistency checking. 
This focus on quality over quantity resulted in a slightly smaller dataset size (1.3K sentences) compared to some previous works like DiMSUM and PARSEME.
However, the evaluation set of CoAM has many more sentences than the gold data of ID10M \cite{tedeschi-etal-2022-id10m}, and is large enough for our purpose of reliably evaluating MWEI systems.

\paragraph{CAIGen}
We observed a usability issue with our annotation interface, CAIGen.
It requires annotators to use separate table rows for each annotated MWE.
Because annotators may forget to use separate rows, which happened in the construction of CoAM, researchers should urge them to confirm their annotations are as intended after finishing each sentence.

\paragraph{MWEI Approaches}

In this study, we experimented with an existing approach, MaW, as well as LLM fine-tuning.
Future work could explore other previously proposed methods, such as BERT-based tagging with BIO-like labels, as introduced by \citet{taslimipoor-etal-2020-mtlb}.


\section*{Ethical Considerations}

All sources of CoAM are permitted at minimum for use in research, as described in \cref{sec:data-sources}.
CoAM itself will be released under the condition that the users do not publish the data on the open web to prevent its leakage to the training data of future models.
All annotators of CoAM consented to the publication of their annotations.
Throughout the dataset construction processes, we found no content harmful to the environment, specific groups of people, or privacy.


\section*{Acknowledgments}

We are grateful for the insightful comments by Yuki Arase,  Shohei Higashiyama, and anonymous reviewers.
To polish the content of this paper, we used ChatGPT\footnote{
    \url{https://chatgpt.com}
} as a writing assistant tool.
Financially, this work was supported by JST SPRING Grant Number JPMJSP2140.


\bibliography{anthology,custom}


\newpage
\quad
\vspace{\textheight-\baselineskip}

\appendix


\begin{table*}[th!]
    \centering
    \small
    \begin{tabular}{
        >{\hspace{-3pt}}
        >{\raggedright\arraybackslash}p{0.07\linewidth}<{\hspace{-3pt}}
        >{\raggedright\arraybackslash}p{0.12\linewidth}<{\hspace{-3pt}}
        >{\raggedright\arraybackslash}p{0.24\linewidth}<{\hspace{-3pt}}
        >{\raggedright\arraybackslash}p{0.15\linewidth}<{\hspace{-3pt}}
        >{\raggedright\arraybackslash}p{0.28\linewidth}<{\hspace{-3pt}}
    }
        \toprule
        Source & Paper & URL & License & Note \\

        \midrule
        News & \citet{glavas-stajner-2013-event, yimam-etal-2018-report} & \url{https://takelab.fer.hr/data/evsimplify/}\newline \url{https://sites.google.com/view/cwisharedtask2018/datasets} & Creative Commons Attribution-NonCommercial-ShareAlike 3.0 & We used the  \texttt{News\_Train} data from the CWI Shared Task 2018 datasets. \\

        \cmidrule(r){1-1} \cmidrule(l){2-5}
        Commentary & \citet{kocmi-etal-2023-findings} & \url{https://data.statmt.org/news-commentary/v18.1/training-monolingual/} & Can be freely used for research purposes. & We used the English part of v18.1. \\
    
        \cmidrule(r){1-1} \cmidrule(l){2-5}
        TED (NAIST-NTT) & \citet{Neubig2014-hv} & \url{https://ahcweb01.naist.jp/old/resource/tedtreebank/} & Creative Commons ShareAlike-Attribution-NonCommercial \\

        \cmidrule(r){1-1} \cmidrule(l){2-5}
        TED (IWSLT) & \citet{Cettolo2012, cettolo-etal-2017-overview} & \url{https://wit3.fbk.eu/2017-01} \url{https://wit3.fbk.eu/2017-01-b} & Creative Commons Attribution-NonCommercial-NoDerivs 3.0 & \\

        \cmidrule(r){1-1} \cmidrule(l){2-5}
        UD & \citet{bies2012english,walsh-etal-2018-constructing} & \url{https://gitlab.com/parseme/parseme_corpus_en} & Creative Commons ShareAlike 4.0 & We used the portion of PARSEME corpus from the English Web Treebank. \\
        \bottomrule
    \end{tabular}
    \caption{Data sources of CoAM.}
    \label{tab:data-sources}
\end{table*}

\section{Construction of CoAM}

\subsection{Data Sources}
\label{sec:data-sources}

See \cref{tab:data-sources}.


\subsection{Notes on Idiomaticity and (Non-)Compositionality}
\label{sec:compositionality}

To judge whether a sequence is an MWE, we focus on (semantic) idiomaticity instead of non-compositionality, although non-compositionality has been considered an inherent property of MWEs in previous studies, such as \citet{tedeschi-etal-2022-id10m}.
To discuss the difference between the two notions, let us consider the expression \textit{spill the beans}.
It is deemed compositional because it can be analyzed as being made up of \textit{spill} in a ``reveal'' sense and the \textit{beans} in a ``secret(s)'' sense, resulting in the overall compositional reading of ``reveal the secret(s)'' \cite{Sag2002}.
Meanwhile, the ``secret(s)'' sense is unique to the expression and cannot be derived from the word \textit{beans}, making the whole expression semantically idiomatic.
We argue that \textit{spill the beans} should be identified as an MWE because the meaning of \textit{beans} depends on the whole expression, and that any idiomatic sequences of this kind should be identified even if they are compositional.






\begin{algorithm*}[ht]
    \small
    \caption{Algorithm of MWE Type Tagging}
    \label{alg:type-tagging}
    
    \begin{algorithmic}[1] 
        \Input{\text{MWE} $\bm{m}$}
        \Output{\text{Type} $t$}

        \ForAll {$w_i \in \bm{m}$}
            \If {$w_i = w^*$}
                \If {PoS of $w_i \in \{\text{NOUN}, \text{PRON}, \text{PROPN}\}$}
                    \If {children of $w_i$ include a relative clause modifier that is verb}
                        \State $t \leftarrow$ \textsc{Verb}  \Comment{E.g., \textit{$\ldots$ the \textbf{price} he \textbf{pays} $\ldots$}}
                    \EndIf

                    \State $t \leftarrow$ \textsc{Noun}  \Comment{E.g., \textit{$\ldots$ born in \textbf{the middle of nowhere} $\ldots$}}

                \ElsIf {PoS of $w_i \in \{\text{VERB}, \text{AUX}\}$}
                    \If {nominal subject of $w_i \in \bm{m}$}
                        \State $t \leftarrow$ \textsc{Clause}  \Comment{E.g., \textit{$\ldots$ \textbf{when it comes to} climate change $\ldots$}}
                    \Else
                        \State $t \leftarrow$ \textsc{Verb}  \Comment{E.g., \textit{$\ldots$ the expression \textbf{stands for} $\ldots$}}
                    \EndIf

                \ElsIf {PoS of $w_i \in \{\text{ADP}, \text{ADJ}, \text{ADV}, \text{CCONJ}, \text{SCONJ}\}$}
                    \State $t \leftarrow$ \textsc{Mod/Conn}  \Comment{E.g., \textit{$\ldots$ for \textbf{at least} two decades $\ldots$}}
                \Else
                    \State $t \leftarrow$ \textsc{Other} (other PoS)  \Comment{E.g., \textit{$\ldots$ \textbf{de facto} target gauge $\ldots$}}
                \EndIf
            \Else
                \State $t \leftarrow$ \textsc{Other} (head not in MWE)  \Comment{E.g., \textit{$\ldots$ , \textbf{and so on} $\ldots$}}
            \EndIf
        \EndFor
        \State \textbf{return} $t$
    \end{algorithmic}
\end{algorithm*}

\subsection{Algorithm for MWE Type Tagging}
\label{sec:appendix-type-tagging}

We tag MWEs in CoAM by the algorithm shown in \cref{alg:type-tagging}.





\begin{table*}[ht]
    \centering
    \small

    \begin{tabular}{lcl}
    \toprule
    MWE Type & Discontinuous (\%) & Example \\
    \midrule
    Noun & \phantom{0}2.0 & \textit{\ldots a \textbf{suicide} car \textbf{bomber} and Taliban militants \ldots} \\
    Verb & 21.8 & \textit{\ldots \textbf{turned} four aircraft \textbf{into} cruise missiles \ldots} \\
    Mod/Conn & \phantom{0}4.8 & \ldots concentration of power \textbf{in} his own \textbf{hands}. \\
    Clause & \phantom{0}0.0 & \\
    Other & 20.8 & \textit{\ldots \textbf{no} more \textbf{than} 2.7\%.} \\
    \bottomrule
    \end{tabular}
    \caption{Ratio of discontinuous MWEs in CoAM by MWE type.}
    \label{tab:discont-ratio}
\end{table*}

\subsection{Statistics Regarding MWE Continuity}
\label{sec:appendix-stats}

Discontinuous MWEs are common in English, and previous work has stressed the importance of identifying them \citep{schneider-etal-2014-discriminative,rohanian-etal-2019-bridging}.
\cref{tab:discont-ratio} shows the ratio of discontinuous MWEs by MWE type.
\textsc{Verb} MWEs have the highest proportion of discontinuous expressions, while \textsc{Mod/Conn}, \textsc{Noun}, and \textsc{Other} MWEs reach lower values.


\newpage
\quad
\newpage
\quad
\vspace{236pt}

\section{Experimental Setup}


\subsection{LLM fine-tuning}
\label{sec:setup-it-appendix}


\begin{table*}[ht]
    \small
    \centering
    \begin{tabular}{lll}
        \toprule
        \multirow[c]{4}{*}{LoRA}
         & $r$ & 64 \\
         & $\alpha$ & 16 \\
         & Dropout & 0.05 \\
         & Target modules & All linear layers \\

        \midrule
        \multirow[c]{6}{*}{Training}
         & Epoch & 3 \\
         & Effective batch size & 32 \\
         & Learning rate & \makecell[l]{1e-4 for Llama-70B \\ 2e-4 for other models} \\
         & Learning rate scheduler & constant \\
         & Optimizer & paged\_adamw\_8bit ($\beta_2$ = 0.999) \\
         & Max grad norm & 0.3 \\
        \bottomrule
    \end{tabular}
    \caption{Hyperparameters for fine-tuning. The epoch was determined based on preliminary experiments, while other parameters are based on \citet{dettmers2023}.}
    \label{tab:hp-it}
\end{table*}


\begin{table*}[ht]
    \centering
    \small
    \begin{tabular}{lp{0.3\textwidth}lp{0.36\textwidth}}
        \toprule
        Name & \multicolumn{3}{l}{Example (Input $\rightarrow$ Output)} \\
        \midrule

        \multirow{-2}{*}{dict\_to\_dict\_list} &
            \begin{verbbox}
{1: 'ACL', 2: 'stands', 
 3: 'for', ...}\end{verbbox}
            \theverbbox &
            \multirow{-2}{*}{$\rightarrow$} &
            \begin{verbbox}
[{'surface': 'stands for',
  'indices': [2, 3]}]\end{verbbox}
            \theverbbox \\

        \cmidrule(r){1-1} \cmidrule(l){2-4}
        \makecell[l]{str\_to\_str\_\\number\_span} &
            \makecell[l]{\ttt{ACL stands for Association} \\ \ttt{for Computational Linguists .}} &
            $\rightarrow$ &
            \makecell[l]{\ttt{ACL <1>stands for</1> Association} \\ \ttt{for Computational Linguists .}} \\

        \cmidrule(r){1-1} \cmidrule(l){2-4}
        \textbf{tsv\_to\_tsv} &
            \makecell[l]{\ttt{ACL\bs n} \\ \ttt{stands\bs n} \\ \ttt{for\bs n} \\ \quad\vdots} & 
            $\rightarrow$ &
            \makecell[l]{\ttt{ACL\bs t\bs n} \\ \ttt{stands\bs t1\bs n} \\ \ttt{for\bs t1\bs n} \\ \quad\vdots} \\

        \bottomrule
    \end{tabular}
    \caption{Input-output formats for fine-tuning. Bold fonts denote the format employed for the main experiments. Suppose the sentence given as the example is \textit{ACL stands for Association for Computational Linguistics}.}
    \label{tab:in-out-format}
\end{table*}


\paragraph{Hyperparameters}
\cref{tab:hp-it} shows hyperparameters used for training and evaluation.
At inference, we perform greedy decoding.


\paragraph{Input-Output Format}
To investigate the optimal input-output format, we perform preliminary experiments.
We prepare three input-output formats described in \cref{tab:in-out-format}.
All the output formats allow us to represent MWEs in any form, including discontinuous or overlapping MWEs.
We train and evaluate Llama-8B and Qwen-7B with CoAM, using each input-output format.
We use the hyperparameters described in \cref{tab:hp-it} and the long MWE definition (same as the main experiments).

As a result, \verb|tsv_to_tsv| turned out to be the only format the models can comply with.
Employing other formats results in the violation of the format despite carefully written instructions.
With \verb|dict_to_dict_list|, the models produce wrong indices in the outputs.
With \verb|str_to_str_number_span|, the models delete spaces before punctuations.


\begin{table*}[ht]
    \centering
    \small
    \begin{tabular}{
        lp{0.9\linewidth}
    }
        \toprule
        Name & Content \\

        \midrule
        \multirow[c]{-13}{*}{Long} & \begin{verbbox}
Here, an MWE is defined as a sequence that satisfies the following three conditions.\n
1. It consists of multiple words that are always realized by the same lexemes. Such words
cannot be replaced without distorting the meaning of the expression or violating language
conventions.\n
2. It displays semantic, lexical, or syntactic idiomaticity. Semantic idiomaticity occurs
when the meaning of an expression cannot be explicitly derived from its components. In
other words, a semantically idiomatic takes on a meaning that is unique to that combination
of words. Lexical idiomaticity occurs when one or more components of an expression are not
used as stand-alone words in standard English. Syntactic idiomaticity occurs when the
grammar of an expression cannot be derived directly from that of its components. For
example, semantically idiomatic MWEs include "break up", the lexically idiomatic include
"to and fro", and the syntactically idiomatic include "long time no see".\n
3. It is not a multi-word named entity, i.e., a specific name of a person, facility, etc.
\end{verbbox}
            \theverbbox \\

        \cmidrule(r){1-1} \cmidrule(l){2-2}
        \multirow[c]{-5}{*}{Short} & \begin{verbbox}
Here, an MWE is defined as a sequence that satisfies the following three conditions.\n
1. It consists of multiple words that are always realized by the same lexemes.\n
2. It is idiomatic, that is, its meaning cannot be explicitly derived from its
components.\n
3. It is not a multi-word named entity, i.e., a specific name of a person, facility, etc.
\end{verbbox}
            \theverbbox \\

        \bottomrule
    \end{tabular}

    \caption{MWE definitions to be included in prompts for fine-tuning.}
    \label{tab:definitions}
\end{table*}

\begin{table*}[ht]
    \centering
    \small

    \begin{tabular}{lcccccccc}
    \toprule
     & & & & & \multicolumn{4}{c}{Recall by MWE Type} \\
    \cmidrule(l){6-9}
     &  & F1 & P & R & Noun & Verb & Mod/Conn & Clause \\
     &  &  &  & (121) & (139) & (111) & (7) \\

    \midrule
    \multirow[c]{2}{*}{Long} & Llama-8B & 24.9$_{\pm 3.7}$ & 92.0$_{\pm 2.0}$ & 14.4$_{\pm 2.4}$ & 3.3$_{\pm 0.0}$ & 19.4$_{\pm 2.9}$ & 21.6$_{\pm 4.8}$ & 0.0$_{\pm 0.0}$ \\
     & Qwen-7B & 48.1$_{\pm 1.0}$ & 60.9$_{\pm 0.8}$ & 39.7$_{\pm 1.0}$ & 28.7$_{\pm 1.3}$ & 52.0$_{\pm 0.8}$ & 39.6$_{\pm 0.9}$ & 0.0$_{\pm 0.0}$ \\

    \midrule
    \multirow[c]{2}{*}{Short} & Llama-8B & 21.0$_{\pm 3.1}$ & 89.9$_{\pm 4.7}$ & 11.9$_{\pm 2.0}$ & 3.3$_{\pm 0.0}$ & 16.3$_{\pm 2.3}$ & 16.8$_{\pm 5.0}$ & 0.0$_{\pm 0.0}$ \\
     & Qwen-7B & 36.7$_{\pm 1.3}$ & 65.7$_{\pm 0.7}$ & 25.5$_{\pm 1.3}$ & 15.4$_{\pm 1.7}$ & 32.6$_{\pm 2.2}$ & 29.7$_{\pm 0.0}$ & 0.0$_{\pm 0.0}$ \\
    \bottomrule
    \end{tabular}

    \caption{
        Ablation results, as mean percentage scores of three runs with random seeds.
        The numbers in parentheses are the MWE counts.
        $\pm$ denotes standard deviation.
    } 
    \label{tab:results-it-short}
\end{table*}

\paragraph{MWE Definition}
\label{para:mwe-def}
To validate the efficacy of the long MWE definition, we perform an ablation study comparing the long definition to the short definition.
They are both a summary of the full definition described in \cref{sec:guidelines}.
As shown in \cref{tab:definitions}, the long definition has 162 words while the short is further contracted to 57 words.
We perform experiments with Llama-8B and Qwen-7B, using the \verb|tsv_to_tsv| format and the hyperparameters described in \cref{tab:hp-it} (same as the main experiments).

\cref{tab:results-it-short} presents the result.
For both models, the long definition achieves higher F1 scores than the short by more than 12 points, indicating the effectiveness of a detailed definition.


\subsection{Evaluation Metrics}
\label{sec:metrics}

We use MWE-based (exact-match) precision, recall, and F1 score \cite{savary-etal-2023-parseme}.
Let $G$ be the set of gold MWEs and $H$ the set of predicted MWEs (hypothesis),
\begin{align}
    \mathrm{Recall} &= |G \cap H|/|G| \\
    \mathrm{Precision} &= |G \cap H|/|H|
\end{align}
where each MWE is represented with the ID of the sentence and the IDs of its constituent tokens. 
F1 is the harmonic mean of precision and recall.



\section{Computational Budgets}
\label{sec:appendix-budgets}

For experiments for MaW (Rule+BiEnc and Rule+DCA), we use a single NVIDIA RTX 2080Ti GPU with 11GB RAM.
Each run of training and testing takes around 8 minutes.
Consequently, the total GPU hours for MaW are estimated to be 0.8 hours. 

For LLM fine-tuning, we use NVIDIA RTX 6000 GPU with 48GB RAM for smaller models (Llama-8B and Qwen-7B) and NVIDIA A100 PCIe with 80GB RAM for larger models (Llama-70B and Qwen-72B).
Each run of training and testing takes around 66 minutes for the smaller models and 588 minutes for the larger models. 
Thus, the total GPU hours for fine-tuning experiments of smaller and larger models are estimated to be 6.6 hours and 58.8 hours, respectively.



\onecolumn

\section{Excerpt of Annotation Guidelines}
\label{sec:guidelines}

\subsection{Definition of MWEs}
\label{sec:appendix-definition}

In our definition, MWEs are idiomatic sequences that (a) consist of multiple words, (b) display semantic, lexical, or syntactic idiomaticity, and (c) are not proper nouns.

\begin{enumerate}[a.]
    \item \textbf{An MWE consists of at least two words that are always realized by the same lexemes}.\footnote{Lexeme is a set of words related through inflection. Words belonging to the same lexeme share a common lemma.} For example, the MWE \textit{"break up"} is always realized by (1) \textit{"break"} or its conjugated form such as \textit{"broke"} and (2) \textit{"up"}. Such words cannot be replaced without distorting the meaning of the expression or violating the language conventions.

    \item \textbf{An MWE displays semantic, lexical, or syntactic idiomaticity}. The semantically idiomatic MWEs include \textit{"break up"}, the lexically idiomatic include \textit{"to and fro"}, and the syntactically idiomatic include \textit{"long time no see"}.
    \begin{itemize}
        \item Semantic idiomaticity occurs when \textbf{the meaning of an expression cannot be derived from its components}. That is, you cannot necessarily infer the meaning of the expression even if you know all the senses of its components. In other words, a semantically idiomatic expression takes on a meaning that is unique to that combination of words.\\
        The inferability differs from one expression to another. The meaning of idiomatic MWEs such as \textit{"kick the bucket"} cannot be inferred from its components at all. The meaning of institutionalized phrases like \textit{"traffic light"} is inferable to some degree. Yet \textit{"traffic light"} is semantically idiomatic because it does not mean any type of light related to traffic but a specific type of light.

        \item Lexical idiomaticity occurs when \textbf{one or more components of an expression are not used as stand-alone words in standard English}. Examples include \textit{"bide one's time"}; \textit{"bide"} rarely appears by itself in today's standard English.

        \item Syntactic idiomaticity occurs when \textbf{the grammatical structure of an expression cannot be derived directly from that of its components}. It constitutes expressions whose grammar seems to go against standard English grammatical conventions. This includes expressions such as \textit{"all of a sudden,"} where \textit{"sudden"} appears in its archaic noun form. 
    \end{itemize}

    \item \textbf{An MWE is not a proper noun, such as the name of a specific person, organization, and so forth.} In this project, do not annotate proper nouns unless they are part of an MWE (like \textit{Pavlov} in \textit{"Pavlov's dog"}). Proper nouns usually start with a capital letter, while exceptions like \textit{"iPhone 15"} exist. Below is the full list of proper nouns in our definition.\footnote{
        The 11 types derive from the named entity types of OntoNotes Release 5.0 \cite{weischedel2013}.
    }

    \begin{table}[ht]
        \centering
        \small
        \begin{tabular}{ll}
            \toprule
            Type & Example \\
            \midrule
            People's names & \textit{Shohei Ohtani} \\
            Nationalities or religious or political groups & \textit{African American, Sunni Muslims} \\
            Facilities & \textit{Narita International Airport, Taipei 101} \\
            Organizations & \textit{Procter \& Gamble, Kyoto University} \\
            Geopolitical entities (GPE) & \textit{Los Angeles, the United Kingdom} \\
            Non-GPE locations & \textit{Mount Fuji, Amazon River} \\
            Products & \textit{Toyota Prius, Samsung Galaxy} \\
            Named events & \textit{World War II, Olympic Games Tokyo 2020} \\
            Works of art & \textit{Norwegian Wood, Bohemian Rhapsody} \\
            Named legal documents & \textit{The Magna Carta} \\
            Named languages & \textit{Middle English, American Sign Language} \\
            \bottomrule
        \end{tabular}
        \label{tab:proper_nouns}
    \end{table}
\end{enumerate}

\subsection{Notes}
\label{sec:appendix-notes}

\subsubsection*{MWEs containing replaceable words}

MWEs are not necessarily made of continuous words. 
Some MWEs contain open slots, that is, words that may be replaced with a large or open class of words. 
In annotation, we do not include open slots in MWEs, as illustrated by the following examples:
\vspace{\topsep}
\begin{compactitem}
    \item \textit{You \textbf{took} me \textbf{by surprise}!}
    \item \textit{\textbf{Pick} me \textbf{up} at the station.}
\end{compactitem}
\vspace{\topsep}

\begin{center}
    \includegraphics[width=0.88\linewidth]{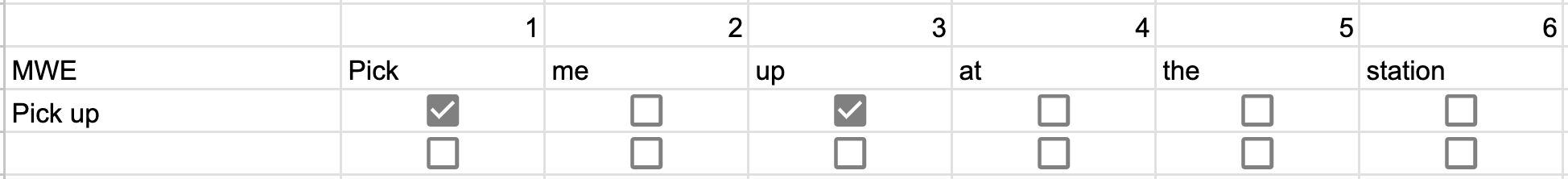}
\end{center}

\noindent Note that if a word is replaceable with only a very small number of alternatives, we consider it as part of the MWE. 
For example, we count the following as different MWEs:
\vspace{\topsep}
\begin{compactitem}
    \item \textit{Their food \textbf{leaves a lot to be desired}.}
    \item \textit{The work \textbf{leaves much to be desired}.}
\end{compactitem}
\vspace{\topsep}

\begin{center}
    \includegraphics[width=1.0\linewidth]{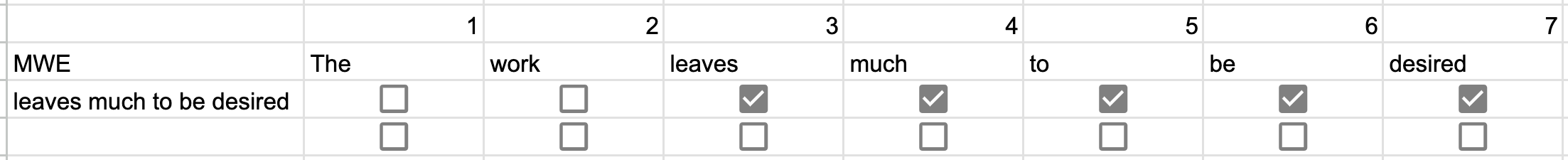}
\end{center}

\subsubsection*{Overlapping MWEs}
\label{sec:appendix-overlapping}

There is a chance that multiple MWEs share the same word. 
For example, \textit{letting} in the following sentence belongs to both \textit{\textbf{letting in}} and \textit{\textbf{letting out}}.

\begin{center}
    \includegraphics[width=1.0\linewidth]{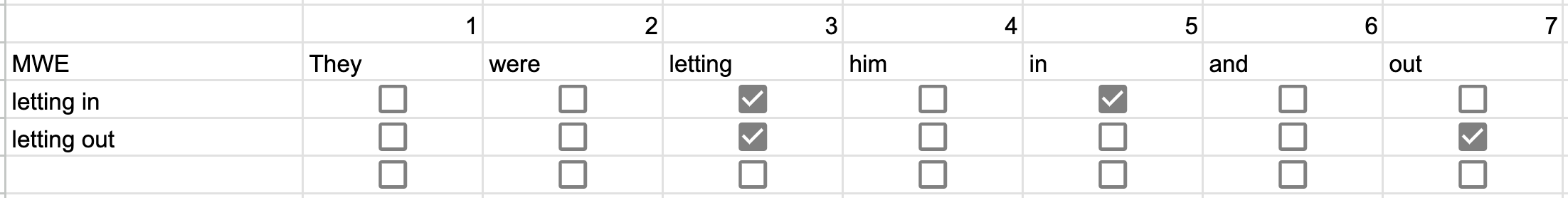}
\end{center}

\subsubsection*{Rearranged MWEs}

There is a chance that the order of component words in an MWE is different from its canonical form, but we still count such rearranged sequences as MWEs. 
In the following example, rearrangement happens in the MWE \textit{\textbf{break} <one’s> \textbf{heart}}. 
As a side note, \textit{my} here is not annotated because it is replaceable.

\begin{center}
    \includegraphics[width=0.7\linewidth]{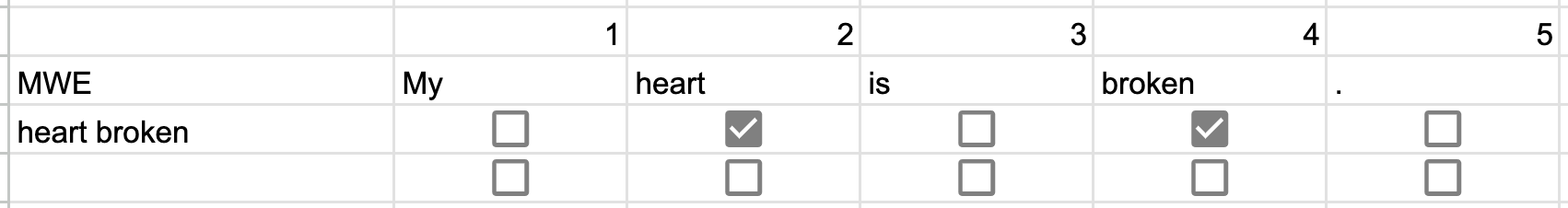}
\end{center}

\subsubsection*{Hyphens}

On our annotation interface, hyphens are treated as words.
When you encounter a hyphen in an MWE, mark the hyphen.
Meanwhile, note that some hyphenated words, e.g., \textit{twenty-one} are not MWEs and should not be annotated.

\begin{center}
    \includegraphics[width=1.0\linewidth]{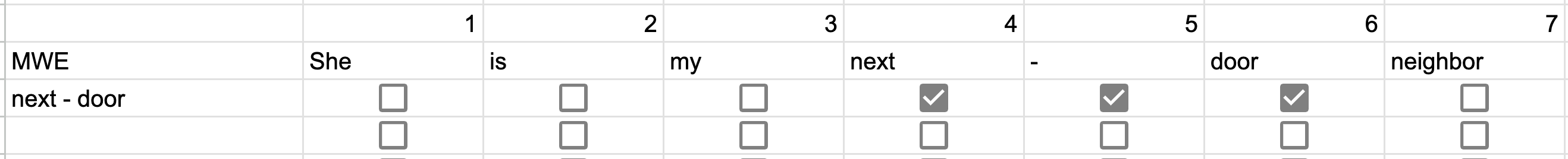}
\end{center}


\end{document}